\documentclass[letterpaper]{article} 
\usepackage{aaai2027}  
\usepackage[hyphens]{url}  
\usepackage{graphicx} 
\usepackage{natbib}  
\usepackage{caption} 
\usepackage{algorithm}
\usepackage{algorithmic}

\usepackage{newfloat}
\usepackage{listings}
\DeclareCaptionStyle{ruled}{labelfont=normalfont,labelsep=colon,strut=off} 
\floatstyle{ruled}
\newfloat{listing}{tb}{lst}{}
\floatname{listing}{Listing}

\usepackage{booktabs}
\usepackage{amssymb}
\usepackage{xcolor}
\usepackage{colortbl}
\definecolor{recipegreen}{RGB}{0,128,0}
\definecolor{recipered}{RGB}{200,0,0}
\definecolor{rowgray}{RGB}{245,245,245}
\usepackage{multirow}
\nocopyright
\title{SAR2Agri: Learning SAR Intensity Representations for Agricultural Monitoring}
\author{
  Moti Rattan Gupta , Anupam Sobti
}
\affiliations{
    Plaksha University, Mohali, Punjab, India

    moti.gupta@plaksha.edu.in, anupam.sobti@plaksha.edu.in
}

\begin{document}

\maketitle

\begin{abstract}
Agricultural monitoring faces unique challenges, arising from the landscape's complex temporal, phenological, and climate dynamics, yet monitoring them is critical for ensuring food security. Synthetic Aperture Radar (SAR) satellites  offer all-weather day-night imaging capability supporting key monitoring tasks including crop type mapping, yield prediction and phenological event detection. Existing multimodal remote sensing foundation models including TerraMind and CopernicusFM learn SAR representations by grounding them in optical imagery using joint encoding and contrastive learning techniques, while SAR-specific foundation models such as SAR-JEPA, SARMAE, and SAR-W-MixMAE primarily focus on target detection, flood mapping, and land cover classification applications.
Recent work has introduced phenology inspired temporal pretext tasks \cite{gupta2026time2agri} with optical imagery which has shown strong performance on agricultural downstream tasks.
In this work, we propose the first self-supervised learning pipeline focused on using only SAR intensity imagery for agricultural applications.
We improve the temporal pretext tasks through masking and curriculum learning to enhance the pretraining pipeline's ability to capture phenological features from SAR. On the SICKLE benchmark, our final model achieves 84.9\% IoU on crop type mapping, outperforming optical baselines (by 15.3 pt) and existing SAR baselines (by 2.2 pt), demonstrating the effectiveness of our proposed pipeline for pretraining SAR intensity encoders for agricultural monitoring.
\end{abstract}


\section{Introduction}
\label{sec:intro}
\begin{figure}[t]
  \centering
  \includegraphics[width=\linewidth]{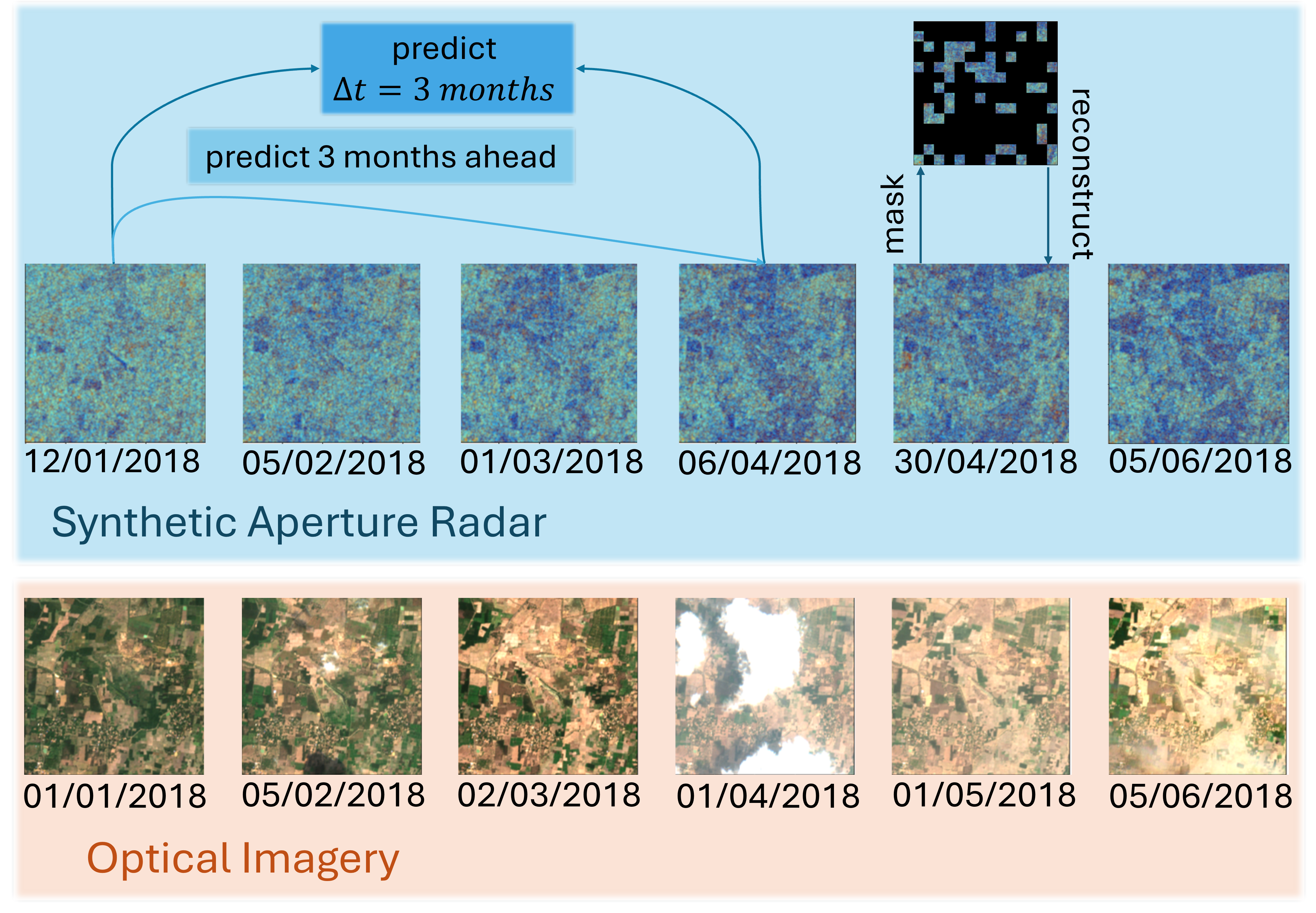}
\caption{\textbf{Pretext-task design space for SAR phenology modeling.}
From a SAR time series, we sample acquisition pairs separated by $\Delta t$ and construct three modular pretext tasks:
(1) predict the time gap $\Delta t$;
(2) forecast the future acquisition from the past;
(3) reconstruct an acquisition from a masked version of itself.
These can be \textbf{combined} (e.g., multi-task learning) into different objectives (Fig~\ref{fig:fig2}).
}
\label{fig:fig1}
\end{figure}

Food security remains a global challenge, with 713--757 million people experiencing hunger in 2023 and 582 million projected to remain undernourished by 2030 \cite{fao2024state}. Remote sensing enables timely and accurate agricultural monitoring, supporting applications including crop type mapping, yield forecasting and phenological event detection. Self-supervised learning (SSL) allows learning representations from large unlabeled optical and radar satellite imagery \cite{lu2025visionfoundationmodelsremote}, which has shown promise in diverse downstream domains including agriculture.

Optical and radar satellites operate in distinct regions of the electromagnetic spectrum and offer complementary views of the landscape. Optical imagery (e.g., Sentinel-2 and Landsat-8) captures spectral properties correlated with vegetation state, but is limited by cloud cover, especially in tropical regions such as Tamil Nadu, India, where the kharif cropping season coincides with persistent monsoon cloud cover.
Synthetic Aperture Radar (SAR) satellites (e.g., Sentinel-1) provide all-weather, day-night imaging by penetrating clouds and capturing structural and pedological properties of vegetation. 
SAR is a complex-valued signal with both amplitude and phase, supporting applications like interferometry and polarimetry \cite{6504845}. We focus on SAR intensity (squared amplitude) which is more widely used in operational monitoring \cite{essd-15-5491-2023} due to its availability on platforms like Google Earth Engine \cite{Gorelick2017GoogleEE} and refer to it as SAR unless otherwise specified.

Existing SAR foundation models (FMs), including SAR-JEPA \cite{LI2024326}, SARMAE \cite{liu2026sarmae}, and SAR-W-MixMAE \cite{caglayan2026sar} are designed for target detection (localizing sparse objects such as vehicles, ships, and aircraft), flood mapping (identifying transient low-backscatter flooded regions), and land cover classification (coarse, scene-level discrimination of croplands, forests, and water bodies). Agricultural monitoring requires dense spatio-temporal scene understanding to form a localized understanding of farm parcels (e.g., crop grown in the parcel) and a broader landscape understanding (e.g., nearby rivers and ponds that shape irrigation and soil moisture). We believe this requires a distinct pretraining pipeline to capture the unique spatio-temporal requirements of agricultural landscapes.

Recently, Time2Agri \cite{gupta2026time2agri} has introduced phenology-aware pretext tasks for optical imagery, arguing that the predictive causality of agricultural land (the natural cycle of sowing, growth and harvest) is an inherent inductive bias that can be leveraged to learn agriculture-relevant representations from optical imagery. Unlike optical imagery, where vegetation dynamics are primarily reflected through chlorophyll-driven spectral changes (captured through high reflectance in NIR and absorption in red), SAR backscatter responds to surface geometry and dielectric properties \cite{ulaby1986microwave} rather than spectral content, with temporal dynamics shaped by canopy volume scattering and soil moisture \cite{steele2017radar}.

Motivated by the distinct properties of SAR, we investigate learning agricultural representations by extending the phenology-inspired pretext tasks to SAR, and studying how to effectively combine them to learn representations that capture both the spatial and temporal dynamics of crop phenology from SAR (Fig~\ref{fig:fig1}). We restrict our scope to regional pretraining and avoid framing this work as a foundation model (FM), and instead focus on improving pretraining efficiency with SAR-only imagery. Regional models have been shown to outperform global counterparts \cite{gupta2026time2agri,tong2026invariantfeaturesglobalcrop}, which is also consistent with broader findings in SSL that suggest domain-aligned pretraining data leads to more effective representations for downstream tasks \cite{al2024pretraining}. We also found that introducing masking as spatial regularization and curriculum learning to combine different temporal objectives leads to further complementary gains over either component alone, when evaluated on the SICKLE benchmark \cite{Sani_2024_WACV} on crop type, yield, and phenological date prediction. 

To the best of our knowledge, this is the first work to design a unimodal SAR pretraining pipeline to generate powerful single SAR intensity image representations for agricultural monitoring. Our key contributions are: \emph{(i) we demonstrate that temporal pretext tasks transfer effectively from optical to SAR, with SAR pretraining substantially outperforming optical for crop type and yield estimation on SICKLE}; \emph{(ii) we show that masking interacts differently with temporal pretext tasks, benefiting dense prediction objectives (predicting future) while degrading global summarization objectives (predicting time differences)}; \emph{(iii) we show that a curriculum over the temporal pretext tasks outperforms both individual and multi-task learning, where we first learn a coarse temporal structure through a global summarization objective before learning more fine-grained temporal dynamics with a dense forecasting objective yielding strong performance on both crop type and yield prediction}; and \emph{(iv) we show that masking and curriculum learning interact non-trivially, with extreme masking in the second stage of the curriculum strengthening rather than degrading the global spatio-temporal prior from the first stage.} 

\begin{figure*}[t]
  \centering
  \includegraphics[width=0.8\textwidth]{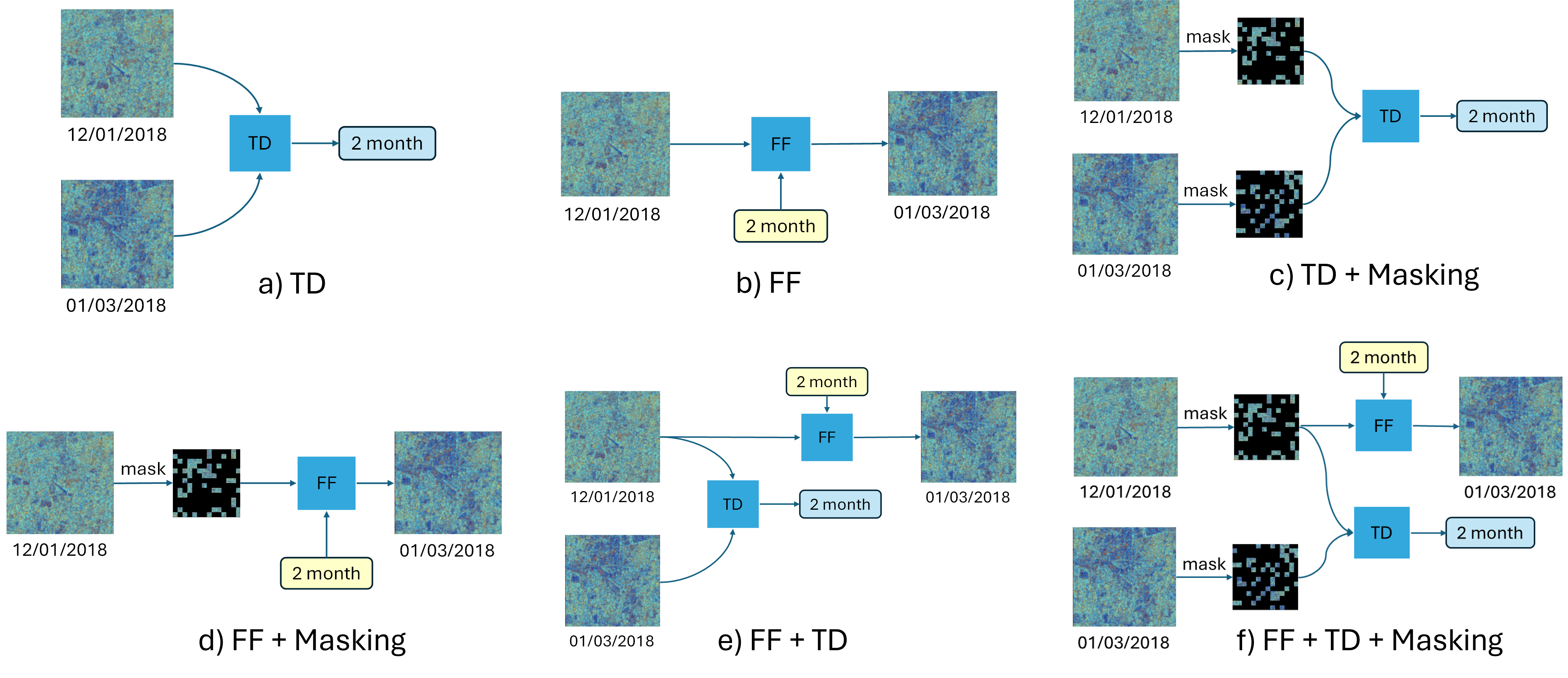}
\caption{Overview of different pretraining strategies.
(a) \textbf{TD:} predict absolute time gap $\Delta t$ from two acquisitions.
(b) \textbf{FF:} predict a future acquisition from a past acquisition.
(c) \textbf{TD + Mask:} predict $\Delta t$ from two independently masked acquisitions.
(d) \textbf{FF + Mask:} predict future acquisition from a masked past acquisition.
(e) \textbf{FF + TD:} jointly predict $\Delta t$ and the future acquisition (both acquisitions for $\Delta t$, past only for forecasting).
(f) \textbf{FF + TD + Mask:} same as (e), with both acquisitions independently masked.
These strategies can be chained as a curriculum — pretrain to convergence on one strategy, then switch to another for continued pretraining (Sec~\ref{sec:curriculum}).}
  \label{fig:fig2}
\end{figure*}

\section{Related Work}

\subsection{SAR in Multimodal Vision RSFMs}
Multimodal Vision RSFMs encode multiple satellite sensors (optical, SAR, and sometimes weather) into a shared representation. CROMA \cite{fuller2023cromaremotesensingrepresentations} aligns masked optical and SAR representations via contrastive learning. DeCUR \cite{wang2024decouplingcommonuniquerepresentations} uses both same-modality and cross-modality contrastive loss to separate common and modality-unique information. DoFA \cite{xiong2024neuralplasticityinspiredmultimodalfoundation} and CopernicusFM \cite{wang2025unifiedcopernicusfoundationmodel} use a modality-specific weight generator to learn a shared encoding using MAE. TerraMind \cite{jakubik2025terramindlargescalegenerativemultimodality} adds a discrete token prediction loss to the existing continuous MAE objective. These models rely on grounding SAR with optical representations via joint encoding \cite{xiong2024neuralplasticityinspiredmultimodalfoundation,wang2025unifiedcopernicusfoundationmodel} or contrastive learning \cite{wang2024decouplingcommonuniquerepresentations,fuller2023cromaremotesensingrepresentations} to learn SAR representations. We focus on learning agricultural representations from single-timestamp SAR imagery in this work.

\subsection{Pretext Tasks for SAR Intensity Imagery}
Early SAR SSL targeted despeckling because of the presence of multiplicative speckle noise. \citet{8519370} showed that cross-acquisition prediction can implicitly despeckle, while SAR2SAR \cite{9399231} despeckles synthetically corrupted images in the log-intensity domain. 
Recent SAR FMs build on MAE \cite{he2022masked} with speckle-related objectives: SAR-JEPA \cite{LI2024326} and SARATR-X \cite{li2025saratrxbuildingfoundationmodel} add gradient feature prediction for speckle-robust features, SAR-W-MixMAE \cite{caglayan2026sar} reweights the loss toward low-backscatter regions, SARMAE \cite{liu2026sarmae} and SUMMIT \cite{DU2025104624} use denoising, with SARMAE additionally adding optical contrastive loss, while SUMMIT adds edge and corner prediction. They perform well on target detection, flood mapping, and land cover classification, but do not focus on capturing temporal or cyclic patterns of crop growth. Our work introduces masking and curriculum over temporal pretext tasks to learn phenology-aware representations from SAR for agricultural monitoring. We avoid speckle-related objectives, since speckle may contain useful information about crop-specific structure.

\subsection{Pretext Tasks for Agricultural Monitoring}

Recently, few agriculture-specific remote sensing models have been proposed, including AgriFM \cite{Li_2026} and Time2Agri \cite{gupta2026time2agri}. AgriFM \cite{Li_2026} is a globally pretrained time-series encoder pretrained on S2, L8, and MODIS with a supervised land cover fraction estimation objective. Time2Agri \cite{gupta2026time2agri} proposes phenology-inspired temporal pretext tasks -- time difference and future frame (forecasting) prediction, which use bitemporal image pairs to pretrain a single-timestamp S2 encoder to learn representations that capture phenological patterns of crop growth. Similarly, the forecasting objective has been proven effective for AEF \cite{brown2025alphaearthfoundationsembeddingfield}, a global pretrained RSFM.
We pretrain single-timestamp SAR encoders for agricultural monitoring to capture the temporal dynamics of crop growth from SAR backscatter, which has fundamentally different properties from optical reflectance.

\section{Methodology}
In this work, we push the boundaries of single-timestamp unimodal SAR  representation learning for agricultural monitoring. We describe our pretraining and evaluation datasets, and the design of our pretraining pipeline, where we design a pretraining curriculum to combine strengths of different temporal pretext tasks and use masking to add further difficulty to the pretraining objective to avoid shortcuts.

\subsection{Datasets}
\paragraph{Pretraining Dataset}
\label{par:pretrainingdata}
Our pretraining pipeline uses temporal pretext tasks \cite{gupta2026time2agri} and requires a dense SAR satellite image time series (SITS). We curate $224\times224$ chips from the S1 RTC \cite{microsoft_planetary_computer} product over Tamil Nadu, using the publicly released locations and timestamps of Time2Agri's \cite{gupta2026time2agri} S2 pretraining dataset (permitting a 21-day time difference). This yields 242,590 chips ($\sim$12b usable pixels) from 6,602 locations, spanning January 2018--March 2021. We apply dB conversion ($10\log_{10}(x)$) to compress dynamic range and convert multiplicative speckle to an additive component following \citet{ulaby2014}.

\paragraph{Evaluation Dataset}
\label{par:evaluationdata}
We use the SICKLE benchmark \cite{Sani_2024_WACV} for evaluation, which provides S1, L8, and S2 SITS over the Cauvery Delta region in Tamil Nadu, India. It has annotations for crop type (1937 train / 227 val. samples), crop yield, sowing, transplanting, and harvesting date (282 train / 37 val. samples). This diverse set of agricultural tasks makes it a suitable benchmark for evaluating our pretraining pipeline. Crop type mapping is a binary segmentation task, measured by intersection-over-union (IoU), while crop yield prediction and phenological date estimation are regression tasks measured by mean-absolute-percentage-error (MAPE). Following \citet{Sani_2024_WACV}, yield prediction uses two settings, one where we have access to SITS from the growing season (in-season), and one where we have access to complete SITS. Date prediction tasks are evaluated on the in-season setting. We use the S1 SITS and the corresponding ground truth labels for our experiments.
\subsection{Setup}
We process our pretraining dataset to form bitemporal image pairs, which are needed for the temporal pretext tasks. Specifically, we define our final pretraining dataset as $S = (X_{t_{i}},X_{t_{j}})$ consisting of bitemporal image pairs
, where $X_{t_{i}}$ and $X_{t_{j}}$ are satellite images at $t_{i}$ and $t_{j}$ with $t_{i} \leq t_{j}$. Each $X_{t_{i}} \in \mathbb{R}^{H \times W \times C}$, where $H$, $W$ and $C$ denote height, width and number of channels respectively, is patchified into $N$ non-overlapping patches, resulting in $X^{patch}_{t_{i}} \in \mathbb{R}^{N \times (P^2  C)}$, where $P$ is the patch size.

We define $f_{\theta}$ as a ViT \cite{dosovitskiy2021imageworth16x16words} encoder which extracts $Z_{t_{i}} \in \mathbb{R}^{(N+1) \times D^{in}}$ from $X^{patch}_{t_{i}} \in \mathbb{R}^{N \times (P^2 C)}$, where $N$ is the number of patches, and define positional encoding \cite{vaswani2023attentionneed} $PE(i,.)$
\begin{equation}
  \label{eq:posemb}
  PE(i,2j) = \sin(\frac{i}{B^{\frac{2j}{D}}})
\end{equation}
\begin{equation}
  PE(i,2j+1) = \cos(\frac{i}{B^{\frac{2j}{D}}})
\end{equation} 
where $i,j$ denote position and feature index, $D$ denotes length of the encoding and $B$ is a constant, which we set to 10000.
Similar to \citet{gupta2026time2agri}, we define time encoding of timestamp $t_{i}$ as $te_{i}$ as
\begin{equation}
  te_{i} = [PE(m_{i});PE(y_{i})]
\end{equation}
where $m_{i},y_{i}$ denote month and year of $t_{i}$, $[;]$ denotes concatenation.

\subsection{Pretext Tasks}

\subsubsection{Time Difference Prediction (TD)} 
TD \cite{gupta2026time2agri} predicts the absolute monthly time difference $|\Delta t| =|t_{2} - t_{1}|$ (Fig~\ref{fig:fig2}(a)) from latent representations $Z_{t_{1}}$ and $Z_{t_{2}}$ by applying a lightweight temporal classifier on top of stacked CLS tokens from both representations
\begin{equation}
  Z_{CLS,t_{1},t_{2}} = [(Z_{t_{1}})_{CLS};(Z_{t_{2}})_{CLS}]
\end{equation}
where $(Z_{t_{i}})_{CLS}$ denote the CLS token of $Z_{t_{i}}$.
The classifier is trained to predict $|\Delta t|$ discretized into $C_{T}$ classes, and optimized using a cross-entropy loss.
\subsubsection{Future Frame Prediction (FF)}
FF \cite{gupta2026time2agri} predicts the future image $X_{t_{2}}$ from the past image $X_{t_{1}}$ of a given location (Fig~\ref{fig:fig2}(b)). FF uses a transformer decoder $tt_\gamma$, called time translator, to predict future latent representation $\hat{Z}_{t_2}$ from the summation of $Z_{t_{1}}$ and concatenation of $te_{1}$, $te_{2}$ and $pe$, and then uses another transformer decoder $g_\phi$ to project the predicted latent to pixel space to get the predicted future frame $\hat{X}_{t_2}$.
\begin{equation}
  \hat{Z}_{t_{2}} = tt_{\gamma}([Z_{t_{1}}] + [pe;te_{1};te_{2}])
\end{equation}
where we set dimension of $pe$, $te_{1}$, and $te_{2}$ to be $D/3$, so their concatenation has the same dimension as $Z_{t_{1}}$.
\begin{equation}
  \hat{X}_{t_{2}} = g_{\phi}(\hat{Z}_{t_{2}})
\end{equation}
The network is optimized using norm-pix-loss \cite{he2022masked} between predicted future $\hat{X}_{t_{2}}$ and ground truth $X_{t_{2}}$.

\subsection{Adding Spatial Objective through Masking}
Inspired by MAE \cite{he2022masked}, we introduce masking to the temporal pretext tasks to further enhance the model's spatial understanding. Specifically, we randomly mask a proportion $p$ of the input image patches by sampling masked indices $M$ and remove them from the input.
\begin{equation}
X^{masked}_{t_{i}} = stack([X^{patch}_{t_{i}}]_{j \notin M})
\end{equation}
where $[X^{patch}_{t_{i}}]_j \in \mathbb{R}^{P^2 C}$ denotes the $j^{th}$ patch of $X^{patch}_{t_{i}}$, and $stack$ denotes the operation of stacking the unmasked patches together, which are then fed into the encoder $f_{\theta}$ to get the masked latent representation $Z^{masked}_{t_{i}}$.

\subsubsection{Masking in TD}
To add masking to TD (Fig~\ref{fig:fig2}(c)), we independently mask both $X_{t_{1}}$ and $X_{t_{2}}$, and use the CLS token of the masked latent representation $Z^{masked}_{t_{i}}$ to predict the time difference.
\subsubsection{Masking in FF}
Since FF is a dense prediction task, similar to MAE \cite{he2022masked}, we only mask the past frame $X_{t_{1}}$ to obtain $Z^{masked}_{t_{1}}$, and use it to predict the future frame $\hat{X}_{t_{2}}$ (Fig~\ref{fig:fig2}(d)). Specifically, we concatenate a learnable mask token $m$, similar to \citet{he2022masked}, to the masked latent representation $Z^{masked}_{t_{1}}$ before feeding it into the time translator $tt_\gamma$ and decoder $g_\phi$ to predict the future frame.

\subsection{Joint Training of Pretext Tasks}
As the temporal pretext tasks capture different aspects of the temporal dynamics, we want to discover how to effectively leverage their complementary strengths, by considering two training strategies:
\subsubsection{Multi-Task Learning}
\label{sec:mtl} 
Multi-Task Learning (MTL) jointly optimizes the model on both pretext tasks by optimizing a weighted sum of the losses of both tasks (Fig~\ref{fig:fig2}(e)-(f)).
\begin{equation}
  L_{MTL} = \lambda L_{FF} + (1-\lambda) L_{TD}
\end{equation}
where $L_{FF}$ and $L_{TD}$ denote the loss of FF and TD respectively, and $\lambda$ is a hyperparameter that controls the relative importance of the two tasks.

\subsubsection{Curriculum Learning}
\label{sec:curriculum}
MTL can be challenging to optimize, and its pareto-optimal solution may not be optimal for downstream tasks \cite{10377704}. Following \citet{10377704}, who show that sequential pretext-task learning can outperform MTL and argue that retaining previous tasks in later stages can retain the strengths of individual tasks, we implement curriculum learning by pretraining on one pretext task (e.g., TD), and then using the resulting model as initialization to jointly (multi-task) train on both that task and an additional one (e.g., FF).

\begin{equation}
  L_{stage_1} = L_{Task_1}
\end{equation}
\begin{equation}
  L_{stage_2} = \lambda L_{Task_1} + (1-\lambda) L_{Task_2}
\end{equation}
where $Task_1$ and $Task_2$ denote the first and second pretext tasks in the curriculum, and $\lambda$ is a hyperparameter that controls the relative importance of the two tasks in the second stage of training.

\section{Experiments}
\label{sec:experiments}
\begin{table*}[t]
  \small
  \centering
  \begin{tabular}{llccccccc}
    \toprule
    & \textbf{Model} & \textbf{Modality} & \textbf{Crop Type} & \textbf{Crop Yield}   & \textbf{Crop Yield}   & \textbf{Sowing}       & \textbf{Transpl.}     & \textbf{Harvest} \\
    &               &                   &                    &                       & \textbf{In-Season}    & \textbf{Date}         & \textbf{Date}         & \textbf{Date} \\
    &               &                   & (IoU\% $\uparrow$) & (MAPE\% $\downarrow$) & (MAPE\% $\downarrow$) & (MAPE\% $\downarrow$) & (MAPE\% $\downarrow$) & (MAPE\% $\downarrow$) \\
    \midrule
    \multirow{5}{*}{\textbf{M. RSFMs}}
    & DoFA           & S1           & 61.215             & 316.616               & 472.536                & 96.316                & 291.242              & 774.963 \\
    & DoFA           & S2           & 58.363             & 36.144                & 34.571                 & 1.685                 & \textit{2.248}       & 7.149 \\
    & CopFM          & S1           & 68.103             & 31.395                & 34.398                 & 2.5443                & 2.834                & \textbf{1.167} \\
    & CopFM          & S2           & 57.229             & 32.334                & 33.706                 & 2.135                 & 2.823                & 6.834 \\
    & TerraMind      & S1           & 75.930             & 33.071	               & 33.390	                & 1.942	                & 3.071	               & 8.563 \\
    \midrule
    \multirow{2}{*}{\textbf{SAR FMs}}
    & SAR-JEPA       & S1           & 76.725             & 33.551                & 34.300                 & 2.433                 & 3.012                & 10.286 \\
    & SARATR-X       & S1           & 61.218             & 38.570                & 33.267                 & 2.491                 & 3.466                & 18.580 \\ 
    & S-W-MMAE       & S1           & 50.791             & 388.010               & 377.840                & 4.364                 & 69.162               & 77.470 \\
    & SARMAE         & S1           & 69.211             & 34.312                & 33.705                 & 2.001                 & 3.140                & 11.174 \\
    \midrule
    \multirow{4}{*}{\textbf{Sup.}}
    & ViT-S          & S1           & 69.185             & \textit{31.213}       & 34.111                 & 2.197                 & 3.007                & 10.328 \\
    & ViT-S          & S2           & 65.523             & 36.357                & 33.687                 & 1.628                 & 2.553                & 6.959 \\
    & UNet3D         & S1           & 82.670             & 31.995                & 34.918                 & \textit{1.460}        & 2.493                & \textit{4.375} \\
    & UNet3D         & S2           & 65.442             & 38.805                & 37.964                 & \textbf{1.248}        & \textbf{1.925}       & 4.760 \\
    \midrule
    \multirow{3}{*}{\textbf{S2 SSL}}
    & FF             & S2           & 69.595             & 36.765                & \textit{32.516}        & 1.582                 & 2.839                & 6.806 \\
    & TD             & S2           & 66.628             & 36.875                & 32.877                 & 1.784                 & 2.499                & 6.595 \\
    & MAE            & S2           & 67.138             & 37.833                & 32.701                 & 1.749                 & 2.341                & 6.520 \\
    \midrule
    \multirow{4}{*}{\textbf{S1 SSL}}
    & FF             & S1           & 82.164             & 31.697                & 33.070                 & 2.108                 & 2.861                & 9.991 \\
    & TD             & S1           & \textit{82.712}    & 35.683                & 33.289                 & 2.173                 & 2.967                & 10.417 \\
    & MAE            & S1           & 76.899             & 35.596                & 32.950                 & 1.859                 & 3.196                & 10.239 \\
    & Ours\footnotemark           
                   & S1           & \textbf{84.871}    & \textbf{30.013}       & \textbf{30.024}        & 2.292                 & 2.905                & 10.509 \\
    \bottomrule
  \end{tabular}
  \caption{Comparison of temporal pretext tasks (FF, TD), MAE \cite{he2022masked}, pretrained multimodal vision RSFMs (M. RSFMs) DoFA \cite{xiong2024neuralplasticityinspiredmultimodalfoundation}, CopernicusFM (CopFM) \cite{wang2025unifiedcopernicusfoundationmodel}, TerraMind \cite{jakubik2025terramindlargescalegenerativemultimodality}, pretrained SAR FMs SAR-JEPA \cite{LI2024326}, SARATR-X \cite{li2025saratrxbuildingfoundationmodel}, SAR-W-MixMAE (S-W-MMAE) \cite{caglayan2026sar}, SARMAE \cite{liu2026sarmae}, supervised baselines (ViT-S, UNet3D) and our proposed method on the SICKLE benchmark, evaluated separately on Sentinel-1 (S1) and Sentinel-2 (S2) splits across six agricultural monitoring tasks. $\uparrow$ indicates higher is better; $\downarrow$ indicates lower is better. \textbf{Bold} and \textit{italic} denote the best and second-best per column.
  Note: Pretext tasks FF, TD and MAE are executed on our pretraining dataset (\ref{par:pretrainingdata}) for S1, or \citet{gupta2026time2agri}'s Tamil Nadu S2 pretraining dataset, while globally pretrained FMs use their publicly available checkpoints.}
  \label{tab:opt_sar_comp}
\end{table*}

\begin{table*}[t]
  \small
  \centering
  \begin{tabular}{llccccccc}
    \toprule
    & \textbf{Model} & \textbf{Mask} & \textbf{Crop Type} & \textbf{Crop Yield}   & \textbf{Crop Yield}   & \textbf{Sowing}       & \textbf{Transpl.}     & \textbf{Harvest} \\
    &                & \textbf{Ratio}&                    &                       & \textbf{In-Season}    & \textbf{Date}         & \textbf{Date}        & \textbf{Date} \\
    &                &               & (IoU\% $\uparrow$) & (MAPE\% $\downarrow$) & (MAPE\% $\downarrow$) & (MAPE\% $\downarrow$) & (MAPE\% $\downarrow$) & (MAPE\% $\downarrow$) \\
    \midrule
    \multirow{2}{*}{\textbf{Without Masking}}
    & FF  & --  & 82.164          & 31.697          & 33.070          & 2.108          & 2.861          & \textbf{9.991} \\
    & TD  & --  & \textit{82.712} & 35.683          & 33.289          & 2.173          & 2.967          & \textit{10.417} \\
    \midrule
    \multirow{4}{*}{\textbf{FF + Mask}}
    & FF  & 25\% & \textbf{83.596} & 32.649          & 34.062          & 1.950          & 3.822          & 12.183 \\
    & FF  & 50\% & 76.366          & 36.603          & 31.948          & 2.051          & 3.184          & 11.062 \\
    & FF  & 75\% & 72.105          & \textit{31.330} & \textit{30.380} & 2.072          & 2.981          & 11.031 \\
    & FF  & 90\% & 81.105          & \textbf{30.920} & 33.882          & 2.142          & 2.912          & 10.57  \\
    \midrule
    \multirow{4}{*}{\textbf{TD + Mask}}
    & TD  & 25\% & 76.818          & 33.074          & 32.583          & \textbf{1.874} & 2.958          & 10.598 \\
    & TD  & 50\% & 77.228          & 32.477          & \textbf{29.765} & \textit{1.936} & \textbf{2.642} & 10.887 \\
    & TD  & 75\% & 76.361          & 33.461          & 33.036          & 2.327          & 3.025          & 11.296 \\
    & TD  & 90\% & 79.173          & 34.126          & 31.750          & 2.215          & \textit{2.683} & 11.971 \\
    \bottomrule
  \end{tabular}
  \caption{Effect of masking ratio $p \in \{0.25, 0.50, 0.75, 0.90\}$ on FF and TD pretraining on the SICKLE benchmark, evaluated against unmasked baselines. $\uparrow$ indicates higher is better; $\downarrow$ indicates lower is better. \textbf{Bold} and \textit{italic} denote the best and second-best per column.}
  \label{tab:mask_ratio}
\end{table*}

\begin{table*}[t]
  \small
  \centering
  \begin{tabular}{lcccccccc}
    \toprule
    & \textbf{Stage I} & \textbf{Stage II} & \textbf{Crop Type} & \textbf{Crop Yield}   & \textbf{Crop Yield}   & \textbf{Sowing}       & \textbf{Transpl.}     & \textbf{Harvest} \\
    &                  &                   &                    &                       & \textbf{In-Season}    & \textbf{Date}         & \textbf{Date}        & \textbf{Date} \\
    &                  &                   & (IoU\% $\uparrow$) & (MAPE\% $\downarrow$) & (MAPE\% $\downarrow$) & (MAPE\% $\downarrow$) & (MAPE\% $\downarrow$) & (MAPE\% $\downarrow$) \\
    \midrule
    \multirow{2}{*}{\textbf{W/o Joint}}
    & FF      & --     & 82.164          & \textit{31.697} & \textit{33.070} & \textit{2.108} & \textit{2.861} & \textbf{9.991} \\
    & TD      & --     & \textit{82.712} & 35.683          & 33.289          & 2.173          & 2.967          & 10.417 \\
    \midrule
    \multirow{1}{*}{\textbf{Multi-Task}}
    & FF + TD & --     & 81.464          & 33.151          & \textbf{31.525} & \textbf{2.104} & \textbf{2.579} & 11.176 \\
    \midrule
    \multirow{2}{*}{\textbf{Curriculum}}
    & FF      & FF+TD   & 81.295          & 34.686          & 33.395          & 2.244          & 2.981          & 10.863 \\
    & TD      & FF+TD   & \textbf{83.705} & \textbf{29.519} & 33.856          & 2.446          & 3.291          & \textit{10.413} \\
    \bottomrule
  \end{tabular}
  \caption{Comparison of joint pretraining strategies on the SICKLE benchmark: multi-task learning (FF$+$TD trained simultaneously), two curriculum orderings (FF$\rightarrow$FF$+$TD and TD$\rightarrow$FF$+$TD), and single-task baselines (FF and TD without joint pretraining). $\uparrow$ indicates higher is better; $\downarrow$ indicates lower is better. \textbf{Bold} and \textit{italic} denote the best and second-best per column.}
  \label{tab:curr_mtl}
\end{table*}

\begin{table*}[t]
  \small
  \centering
  \begin{tabular}{lccccccc}
    \toprule
    & \textbf{Stage II}   & \textbf{Crop Type} & \textbf{Crop Yield}   & \textbf{Crop Yield}   & \textbf{Sowing}       & \textbf{Transpl.}     & \textbf{Harvest} \\
    & \textbf{Mask Ratio} &                    &                       & \textbf{In-Season}    & \textbf{Date}         & \textbf{Date}         & \textbf{Date} \\
    &                     & (IoU\% $\uparrow$) & (MAPE\% $\downarrow$) & (MAPE\% $\downarrow$) & (MAPE\% $\downarrow$) & (MAPE\% $\downarrow$) & (MAPE\% $\downarrow$) \\
    \midrule
    \multirow{1}{*}{\textbf{Curriculum W/o Mask}}
    & N/A     & \textit{83.705}  & \textbf{29.519} & 33.856          & 2.446          & 3.291          & \textbf{10.413} \\
    \midrule
    \multirow{3}{*}{\textbf{Curriculum + Mask}}
    & 50\% 	  & 77.838	        & 32.245          & 33.889          & 2.393          & \textbf{2.643} & 11.444 \\
    & 75\%	  & 76.723          & 30.424          & \textit{30.378} & \textbf{2.119} & \textit{2.844} & 10.728 \\
    & 90\%	  & \textbf{84.871} & \textit{30.013} & \textbf{30.024} & \textit{2.292} & 2.905          & \textit{10.509} \\
    \bottomrule
  \end{tabular}
  \caption{Effect of Stage~II masking ratio on the TD$\rightarrow$FF$+$TD curriculum on the SICKLE benchmark, with mask ratios $p \in \{0.50, 0.75, 0.90\}$. The unmasked TD$\rightarrow$FF$+$TD curriculum is included as reference. $\uparrow$ indicates higher is better; $\downarrow$ indicates lower is better. \textbf{Bold} and \textit{italic} denote the best and second-best per column.
  Note: $p = 0.25$ is excluded due to out-of-memory (OOM) during pretraining.}
  \label{tab:curr_mtl_masking}
\end{table*}

\begin{table*}[t]
  \small
  \centering
  \begin{tabular}{clcccc}
    \toprule
    \textbf{Step} & \textbf{Configuration}
      & \textbf{Crop Type}     & \textbf{$\Delta$ Type}
      & \textbf{Crop Yield}    & \textbf{$\Delta$ Yield} \\
                  &
      & (IoU\%\,$\uparrow$)    & (IoU\%)
      & (MAPE\%\,$\downarrow$) & (MAPE\%) \\
    \midrule
    1 & TD
      & 82.712
      & ---
      & 35.683
      & --- \\
    2 & FF\,$+$\,TD
      & 81.464
      & \textcolor{recipered}{$\blacktriangledown$\,1.248}
      & 33.151
      & \textcolor{recipegreen}{$\blacktriangle$\,2.532} \\
    3 & TD\,$\rightarrow$\,FF\,$+$\,TD
      & \textit{83.705}
      & \textcolor{recipegreen}{$\blacktriangle$\,2.241}
      & \textbf{29.519}
      & \textcolor{recipegreen}{$\blacktriangle$\,3.632} \\
    4 & TD\,$\rightarrow$\,FF\,$+$\,TD + 90\% Mask
      & \textbf{84.871}
      & \textcolor{recipegreen}{$\blacktriangle$\,1.166}
      & \textit{30.013}
      & \textcolor{recipered}{$\blacktriangledown$\,0.494} \\
    \bottomrule
  \end{tabular}
  \caption{Progressive ablation of pretraining components on the SICKLE benchmark. Each step adds one component over the previous row; $\Delta$ denotes the absolute change relative to the immediately preceding step. \textcolor{recipegreen}{$\blacktriangle$} and \textcolor{recipered}{$\blacktriangledown$} indicate improvement and degradation respectively. \textbf{Bold} and \textit{italic} denote the best and second-best per column.
  }
  \label{tab:ablation}
\end{table*}

We investigate: (i) do temporal pretext tasks designed for optical imagery remain effective when applied to SAR; (ii) does masking further improve representations learned via temporal pretext tasks; (iii) among multi-task learning and curriculum learning, which can better leverage the complementary strengths of the temporal pretext tasks; and (iv) would adding masking to the curriculum learning strategy further improve the representation.

\subsection{Experimental Setup}

\subsubsection{Baselines}
\label{sec:baselines}
We consider the following baselines for comparison: a supervised ViT-S and UNet3D trained from scratch, a self-supervised MAE \cite{he2022masked} pretrained on single-frame reconstruction with 75\% masking, optical baselines from \citet{gupta2026time2agri} pretrained on S2 SITS (FF, TD, and MAE on Tamil Nadu). These baselines allow us to evaluate the benefits of temporal pretext tasks for SAR, and to compare against pretraining on optical data. We also include globally pretrained RSFMs, namely DoFA \cite{xiong2024neuralplasticityinspiredmultimodalfoundation}, CopernicusFM \cite{wang2025unifiedcopernicusfoundationmodel} and TerraMind \cite{jakubik2025terramindlargescalegenerativemultimodality} due to their strong performance in benchmarks \cite{jakubik2025terramindlargescalegenerativemultimodality}, and global SAR FMs, SAR-JEPA \cite{LI2024326}, SARATR-X \cite{li2025saratrxbuildingfoundationmodel}, SAR-W-MixMAE \cite{caglayan2026sar} and SARMAE \cite{liu2026sarmae}, using their publicly available checkpoints and evaluate their performance on S1 split of the SICKLE benchmark.
\footnotetext{Our proposed pipeline: TD$\rightarrow$TD+FF+90\% Mask (Sec~\ref{par:curr_mask_comp})}

\subsubsection{Implementation Details}
We use a ViT-S encoder for pretraining, with a 3-layer MLP for TD, and a 2-layer Time Translator + 4-layer Transformer decoder for FF. Bitemporal pairs are sampled with a max gap of 3 months. All experiments use VV and VH polarizations as the input channels. We do not use VH/VV ratio as an additional channel, due to its reduced performance in supervised settings \cite{DANDRIMONT2021112708}. 
We use a UPerNet \cite{xiao2018unifiedperceptualparsingscene} segmentation head for all ViT backbones. For pretrained multimodal vision RSFMs and SAR FMs, we specifically use their Base variants for comparison, and adjust the input channels for UPerNet segmentation head \footnote{Please refer to supplementary material for more details.}. 

\subsection{Results}

\subsubsection{Do Optical Temporal Pretext Tasks Transfer to SAR?}
\label{sec:exp_sar_opt_comp}

To test whether optical temporal pretext tasks transfer to SAR, we pretrain ViT-S on S2 and S1 with FF, TD, and MAE, and evaluate on the corresponding SICKLE splits, and compare against the baselines described in Sec~\ref{sec:baselines}.

\paragraph{\textit{Temporal Pretext Tasks Transfer to SAR:}}
Table~\ref{tab:opt_sar_comp} shows that FF (82.2\%) and TD (82.7\%) pretrained on S1 outperform DoFA (61.2\%), CopernicusFM (68.1\%), TerraMind (75.9\%), SAR-JEPA (76.7\%), SARATR-X (61.2\%), SAR-W-MixMAE (50.8\%), SARMAE (69.2\%), MAE (76.9\%) pretrained on our S1 pretraining data, supervised ViT-S (69.2\%), and remain competitive with UNet3D (82.7\%) on crop type mapping on S1. Similarly, FF (31.697\%) remains competitive on yield estimation and date prediction tasks, with exception of CopernicusFM on harvest prediction (1.167\% MAPE), possibly due to inclusion of atmospheric products (Sentinel-5P) in its pretraining. The overall performance suggests that temporal pretext tasks can learn temporal dynamics from SAR as well. The performance gain switching from MAE to FF for crop type is larger for S1 (5 pt\footnote{pt denotes the change between two metric values in \% points.}) than for S2 (2.5 pt), suggesting that the temporal structure captured by FF is particularly beneficial for SAR data.

\paragraph{\textit{SAR Pretraining Outperforms Optical Pretraining for Crop Type and Yield:}}
Comparing S1 and S2 pretrained models, we observe S1 outperforms S2 on FF (by 12.5 pt), TD (15.5 pt), and MAE (10 pt) on crop type, and similarly on yield though with some degradation on date tasks (3.1 pt--3.9 pt for harvest). This suggests date prediction benefits more from visible maturity than structural information from SAR. The overall gain in crop type (16.1 pt on TD) and yield (5 pt on FF) suggests S1 pretraining is more effective for crop type and yield prediction. CopernicusFM shows a similar S1 advantage on crop type, while DoFA's S1 performance substantially degrades on most tasks, suggesting its SAR representations may not transfer well for agricultural monitoring. 

\subsubsection{Can Masking Improve Pretraining With FF and TD?}
\label{sec:exp_masking}

To investigate the impact of masking of input patches on temporal pretext tasks, we sweep mask ratios $p \in \{0.25, 0.5, 0.75, 0.9\}$ for both FF and TD, and evaluate them on SICKLE.

\paragraph{\textit{Masking Improves FF but Degrades TD:}}
\label{par:ff_td_mask_comp}
Table~\ref{tab:mask_ratio} shows a low mask ratio ($p{=}25\%$) on FF improves crop type (82.2\% vs 83.6\%) with slight yield degradation (0.95 pt) while TD degrades at all ratios (76.4--79.2\%). Date tasks experience minimal loss across ratios, except harvest performance which increases with increasing mask ratios on FF but degrades on TD. The overall pattern is consistent with task design, where a dense prediction objective (FF) can benefit from mild masking which encourages learning spatial associations, while a CLS-based global summarization task (TD) is more sensitive to missing spatial context, making masking less suitable.

\paragraph{\textit{What Is the Optimal Masking Ratio for FF?}}
\label{par:mask_tradeoff}
Table~\ref{tab:mask_ratio} shows 25\% masking leads to good crop type (83.6\%) but poor yield (32.7\%) performance, while a 90\% masking leads to good yield (31\%) but poor crop type (81.1\%) performance, despite 25\% showing overall better pretraining loss (0.69 vs 0.93). This indicates general useful representation across tasks could not be learnt, which we address through curriculum learning (Sec~\ref{sec:exp_curriculum}).

\subsubsection{Combining FF and TD}
\label{sec:exp_curriculum}

We now investigate how to effectively combine FF and TD to leverage their complementary strengths, by comparing MTL (Sec~\ref{sec:mtl}) and curriculum learning (Sec~\ref{sec:curriculum}). We consider both curricula: FF followed by FF+TD (FF$\rightarrow$FF+TD) and TD followed by FF+TD (TD$\rightarrow$FF+TD) to understand role of task order.

\paragraph{\textit{Curriculum vs Multi-Task Learning:}}
Table~\ref{tab:curr_mtl} shows that a well-chosen curriculum (TD$\rightarrow$FF+TD) can outperform MTL, achieving better crop type (83.7\% vs.\ 81.3\%) and overall yield (29.5\% vs.\ 33.2\%) than FF+TD trained jointly, while remaining competitive on date tasks.

\paragraph{\textit{Does Task Order Matter for Curriculum Learning?}}
TD$\rightarrow$FF+TD outperforms FF$\rightarrow$FF+TD on crop type (2.4 pt) and yield (5.2 pt), improving over the single-task baselines and indicating that task order is important. This is consistent with the intuition that TD, by summarizing global temporal context, can provide a rich spatio-temporal prior for the subsequent FF stage to learn more richer spatial associations, while the reverse (FF$\rightarrow$FF+TD) might not provide the same benefits. We use the TD$\rightarrow$FF+TD curricula as the basis for our later experiments.

\paragraph{\textit{Does Masking Further Improve the Curriculum?}}
\label{par:curr_mask_comp}
On TD$\rightarrow$FF+TD, we investigate whether adding masking to Stage~II can further improve performance. As shown in Table~\ref{tab:curr_mtl_masking}, only 90\% masking improves crop type (84.9\% vs.\ 83.7\% IoU) and in-season yield (30.024\% vs 33.856\%) over the unmasked curriculum, while 50\% and 75\% masking significantly degrade crop type performance (77.8\% and 76.7\% respectively). This is accompanied by minimal loss in sowing (0.173 pt) and transplanting (0.262 pt), and a slight gain in harvest (0.935 pt) performance across masking ratios. The effectiveness of extreme masking (90\%) can be explained by understanding the role of spatial masking on the Stage~I TD initialization. The initial TD pretraining provides a rich spatio-temporal prior which can weaken on addition of easier spatial tasks (50\% and 75\% masking), as it allows the model to rely on local spatial cues for reconstruction. In contrast, extreme masking (90\%) can reduce this spatial reliance by forcing the model to improve this existing prior by reconstructing heavily occluded inputs, improving performance on both crop type and yield prediction.

\paragraph{\textit{Overall Findings}}
\label{sec:exp_summary}
Table~\ref{tab:ablation} summarizes the progressive contribution of each component: naive MTL is suboptimal (crop type degrades by $1.2$ pt with yield gain of $2.5$ pt), curriculum learning recovers and exceeds single-task performance (83.7\% IoU, 29.5\% MAPE), and adding 90\% masking further improves crop type to 84.9\% IoU at a marginal 0.5 pt yield cost — confirming that pretext-task structure and order affect downstream performance. Our final pipeline (TD $\rightarrow$ FF+TD + 90\% mask) outperforms optical pretraining by 15.3 pt, supervised baselines by 2.2 pt, and even globally trained multimodal vision RSFMs by 9 pt and SAR-specific FMs by 8.1 pt on crop type, as well as yield. We see slight deterioration on sowing (0.7 pt), transplating (0.6 pt) and harvest (4 pt) prediction compared to optical pretraining, hinting at slight spatial learning bias of the pipeline (Table~\ref{tab:opt_sar_comp}).

\section{Conclusion}
We develop a state-of-the-art self-supervised pretraining pipeline for agricultural monitoring using only SAR intensity imagery. We use temporal pretext tasks (FF and TD) designed for optical imagery but add a curriculum (TD$\rightarrow$TD+FF) and masking (90\%) — showing that temporal pretext tasks transfer to SAR intensity, that masking benefits dense (FF) but harms global-summarization (TD) objectives, and that a TD-learned global prior provides better initialization for further curriculum pretraining than joint training alone. On regional pretraining over Tamil Nadu, we show it consistently outperforms optical pretraining (by 15.3 pt), supervised baselines (by 2.2 pt) and even globally pretrained multimodal vision RSFMs (by 9 pt) and SAR FMs (by 8.1 pt) on crop type, as well as yield estimation on the SICKLE benchmark, with slight deterioration on date tasks compared to optical pretraining. 
SAR2Agri establishes phenology-aware pretraining as a promising approach for agricultural monitoring using SAR intensity. In the future, we plan on creating a benchmark across a diverse set of regions to understand the regional variances of these representations.

\section{Acknowledgements}
We are grateful to Dr. Shashank Tamaskar for his valuable feedback during various stages of this work. This research was supported by the Plaksha University Startup Research Grant and by GPU compute credits provided by Jarvis Labs.

\clearpage
\appendix


\section{Pretraining Dataset Construction}
\label{sec:pretrainingdata}
\begin{table*}[t]
\centering
\begin{tabular}{lccc}
\toprule
\textbf{Component} & \textbf{Pretraining} & \textbf{Downstream} & \textbf{Downstream A100} \\
\midrule
Hardware  & RTX A6000 (48GB)                 & RTX A5000 (24GB) & A100 (80GB) \\
OS        & Ubuntu 24.04                     & Ubuntu 22.04     & Ubuntu 22.04  \\
Framework & PyTorch 2.10.0 + Lightning 2.6.1 & PyTorch 2.6.0    & PyTorch 2.11.0  \\
CUDA      & 13.1                             & 12.2             & 13.0 \\
Training  & Single GPU                       & Single GPU       & Single GPU \\
\bottomrule
\end{tabular}
\caption{\textbf{Experimental Environment}}
\label{tab:env}
\end{table*}

\begin{table*}[t]
\centering
\begin{tabular}{lccccccc}
\toprule
\textbf{Parameter} & \textbf{FF}   & \textbf{TD}    & \textbf{FF+TD} & \textbf{MAE} & \textbf{FF+Mask} & \textbf{TD+Mask} & \textbf{FF+TD+Mask}  \\
\midrule
Learning Rate     & 0.0001         & 0.0009         & 0.0009         & 0.001 & 0.0001 & 0.0001 & 0.0001\\
Batch Size        & 512            & 256            & 256            & 1024 & 512 & 512 & 512 \\
Image Resolution  & 224$\times$224 & 224$\times$224 & 224$\times$224 & 224$\times$224 & 224$\times$224 & 224$\times$224 & 224$\times$224 \\
Warmup Epochs     & 10             & 10             & 10             & 10 & 10 & 10 & 10 \\
Max Epochs        & 100            & 100            & 100            & 100 & 100 & 100 & 100 \\
Early Stop Patience & 5            & 5              & 5              & 5 & 5 & 5 & 5 \\
Optimizer         & AdamW          & AdamW          & AdamW          & AdamW & AdamW & AdamW & AdamW \\
LR Schedule       & Cosine         & Cosine         & Cosine         & Cosine & Cosine & Cosine & Cosine \\
Weight Decay      & 0.01           & 0.01           & 0.01           & 0.01 & 0.01 & 0.01 & 0.01 \\
Loss Function     & NPE            & CE             & weighted       & NPE & NPE & CE & weighted\\
                  &                &                & CE \& NPE      & & & & CE \& NPE\\  
$\lambda$ (for Joint Training)& --    & --             & 0.5            & -- & -- & -- & 0.5 \\
\bottomrule
\end{tabular}
\caption{\textbf{Pretraining Hyperparameters}: This table describes the hyperparameters for FF, TD, FF+TD, MAE, TD+Masking (TD+Mask), FF+Masking (FF+Mask) and FF+TD+Masking (FF+TD+Masking) pretraining. Note: the hyperparameters for curriculum learning strategies (FF$\rightarrow$FF+TD, TD$\rightarrow$FF+TD and TD$\rightarrow$FF+TD+Masking) use the values as the mentioned depending on the stage in the curriculum.}
\label{tab:pretrain_hparams}
\end{table*}

\begin{table*}[t]
\centering
\begin{tabular}{lcccccc}
\toprule
\textbf{Parameter}        & \textbf{ViT-S} & \textbf{DoFA}   & \textbf{SAR-JEPA}& \textbf{SARATR-X} & \textbf{SAR-W-MixMAE} & \textbf{UNet} \\
                          & \textbf{Backbones}& \textbf{CopFM}  &                & & & \\
                          &                & \textbf{TerraMind}&              & & & \\
                          &                & \textbf{SARMAE}   &              & & & \\
\midrule
Learning Rate             & 0.001          & 0.001          & 0.001          & 0.001  & 0.001 & 0.001 \\
Batch Size                & 192            & 64             & 128            & 64 & 192 & 192 \\
Image Resolution          & 64$\times$64   & 64$\times$64   & 64$\times$64   & 64$\times$64 & 64$\times$64 & 64$\times$64 \\
Max Epochs                & 100            & 100            & 100            & 100 & 100 & 100 \\
Optimizer                 & Adam           & Adam           & Adam           & Adam & Adam & Adam \\
Loss Function (Crop Type) & CE             & CE             & CE             & CE & CE & CE \\
Loss Function (Rest)      & RMSE           & RMSE           & RMSE           & RMSE & RMSE & RMSE \\
GPU                       & A5000          & A5000          & A5000          & A100 (80GB) & A100 (80GB) & A5000 \\  
\bottomrule
\end{tabular}
\caption{\textbf{Downstream Hyperparameters for Evaluation on the SICKLE benchmark}: This table describes the hyperparameters for downstream evaluation on the SICKLE benchmark for crop type mapping, yield estimation, and phenological date prediction. Note: the same hyperparameters are used across all three downstream tasks except for the loss function, where crop type mapping uses cross-entropy (CE) loss while yield estimation and phenological date prediction use root mean squared error (RMSE) loss. SARATR-X and SAR-W-MixMAE use A100 (80GB) due to their high compute requirement.}
\label{tab:downstream_hparams}
\end{table*}

\begin{table*}[t]
\centering

\begin{tabular}{@{}c@{\hspace{1.5em}}c@{}}

\begin{tabular}[t]{@{}ll@{}}
\toprule
\textbf{Component} & \textbf{Specification} \\
\midrule
Architecture        & ViT (Small) \\
Patch Size          & $16 \times 16$ \\
Embedding Dimension & 384 \\
Number of Layers    & 12 \\
Number of Heads     & 12 \\
MLP Ratio           & 4.0 \\
Input Channels      & 2 (VV + VH) \\
\bottomrule\\[3pt]
\multicolumn{2}{c}{\textbf{(a) Backbone Configuration}} 
\end{tabular}

&

\begin{tabular}[t]{@{}lllccc@{}}
\toprule
\textbf{Method} &
\textbf{Component} &
\textbf{Architecture} &
\textbf{Layers} &
\textbf{Heads} &
\textbf{Dim} \\
\midrule
FF
  & Time Translator & Transformer & 2 & 12 & 384 \\
  & Decoder         & Transformer & 4 & 4  & 192  \\
\addlinespace
TD
  & Classifier      & MLP         & 3 & -- & 192 \\
\addlinespace
FF + TD
  & Time Translator & Transformer & 2 & 12 & 384 \\
  & Decoder         & Transformer & 4 & 4  & 192 \\
  & Classifier      & MLP         & 3 & -- & 192 \\
\addlinespace
MAE
  & Decoder         & Transformer & 6 & 6  & 192 \\
\bottomrule\\[3pt]
\multicolumn{6}{c}{\textbf{(b) Task-Specific Components}} 
\end{tabular}

\end{tabular}

\caption{\textbf{Architecture configurations:}
(a) This table describes the ViT-S \cite{dosovitskiy2021imageworth16x16words} backbone architecture used for all pretraining methods. 
(b) This table describes the architectural details for the task-specific components for each method.}
\label{tab:architecture_details}
\end{table*}

To pretrain our SAR encoder, we curate a large satellite image time series (SITS) of S1 intensity imagery over Tamil Nadu, India. We illustrate the pipeline for curating the S1 pretraining dataset as follows (Fig~\ref{fig:pretrain_data}):
\begin{enumerate}
  \item \textbf{Download Reference Locations \& Timestamps} We first download the reference locations and timestamps of S2 imagery from the regional pretraining dataset of Time2Agri \cite{gupta2026time2agri}, which we use as a reference to query for S1 imagery, which are chips of shape $224 \times 224$ spanning 6602 unique spatial locations with temporal coverage from January 2018 to March 2021 at a monthly cadence. We use the S2 locations and timestamps as reference for querying S1 imagery to ensure spatial alignment between the two modalities, to ensure a fair comparison on the role of modality for learning agricultural representations, and to enable potential future work on multimodal pretraining by learning from the same locations and timestamps across both modalities.

  \item \textbf{Query Sentinel-1 RTC Collection} For each chip boundary and S2 timestamp, we query the Sentinel-1 RTC collection on the Microsoft Planetary Computer \cite{microsoft_open_source_2022_7261897} for available S1 acquisitions over that location. We use the RTC product because it is already terrain corrected and projected to a regular grid, which allows us to directly apply the pretext tasks designed for regular grid data (e.g., optical imagery).
  
  \item \textbf{Filter By Temporal Proximity}
  For the available S1 acquisitions from the previous step, we filter them by temporal proximity to the S2 timestamp, where we select an S1 acquisition only if it is within a 21-day window of the S2 timestamp. If multiple S1 acquisitions are available within the 21-day window, we select the one closest in time to the S2 timestamp, and if no S1 acquisition is available within the 21-day window, we skip that month for that location. We allow a time difference of 21 days to ensure sufficient temporal coverage for the pretraining dataset while ensuring similar phenological state between the S1 and S2 acquisitions. We repeat this process for each monthly S2 timestamp for each location. This results in a total of 242,590 S1 chips across 6602 locations.
  
  \item \textbf{Select and Export to Zarr.} 
  For each chip location, we obtain the satellite image time series (SITS) of the filtered S1 acquisitions, which have two channels corresponding to the VV and VH polarizations, and export them to a Zarr store.

\begin{figure}[H]
  \centering
  \includegraphics[width=0.80\linewidth]{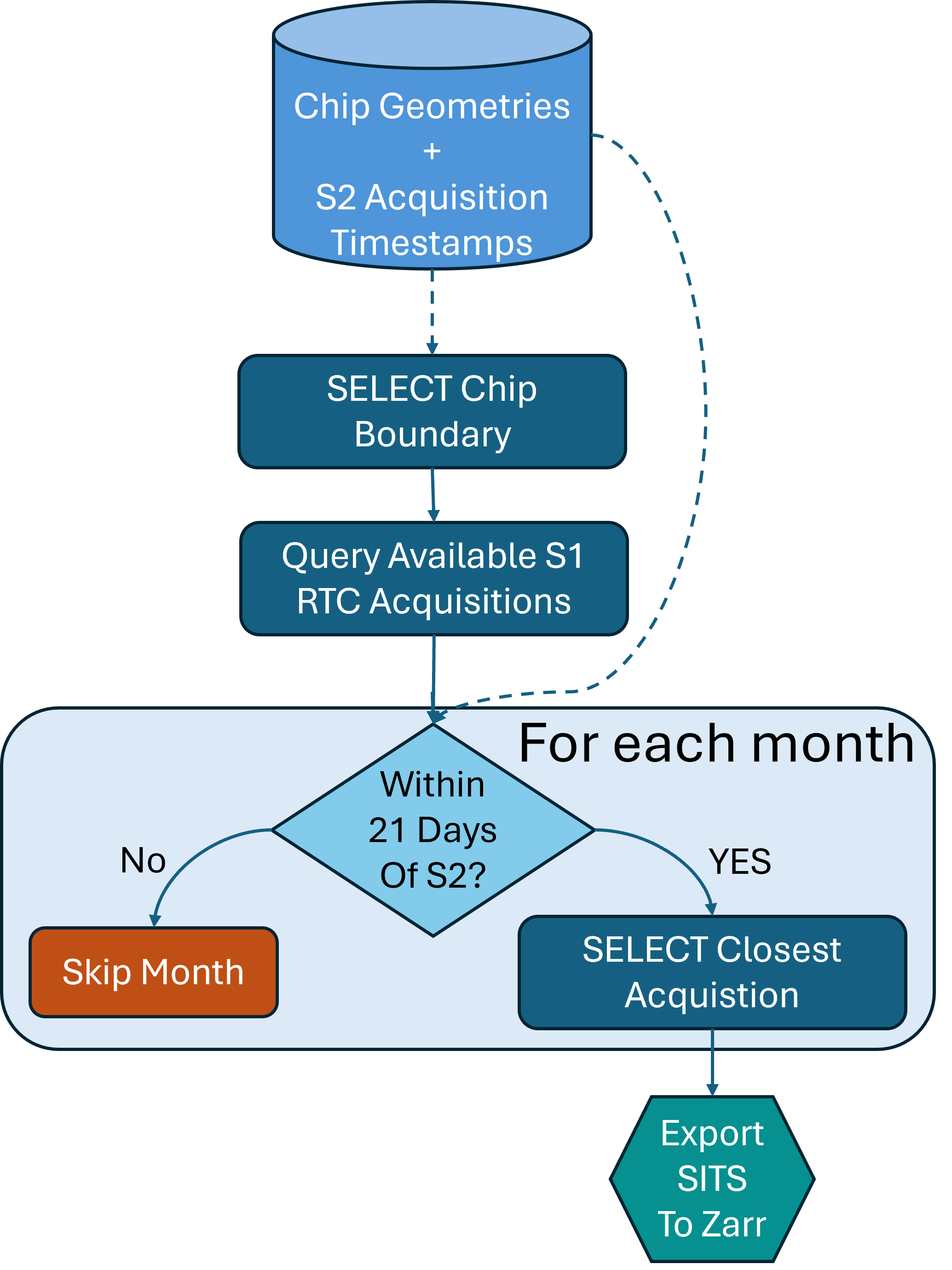}
  \caption{This figure shows the pipeline for constructing the S1 pretraining dataset for Tamil Nadu. We use the publicly released locations and timestamps of S2 imagery from the regional pretraining dataset of Time2Agri \cite{gupta2026time2agri} to sample S1 chips ensuring spatial alignment with an allowed time difference of 21 days. The chips are of the shape $224 \times 224$ and contain 2 channels corresponding to the VV and VH polarizations of S1.}
  \label{fig:pretrain_data}
\end{figure}
\end{enumerate}

\section{Further Investigation on Decline of Harvest Prediction Performance}

\subsection{Role of Modality}
\begin{figure}[H]
  \centering
  \includegraphics[width=\linewidth]{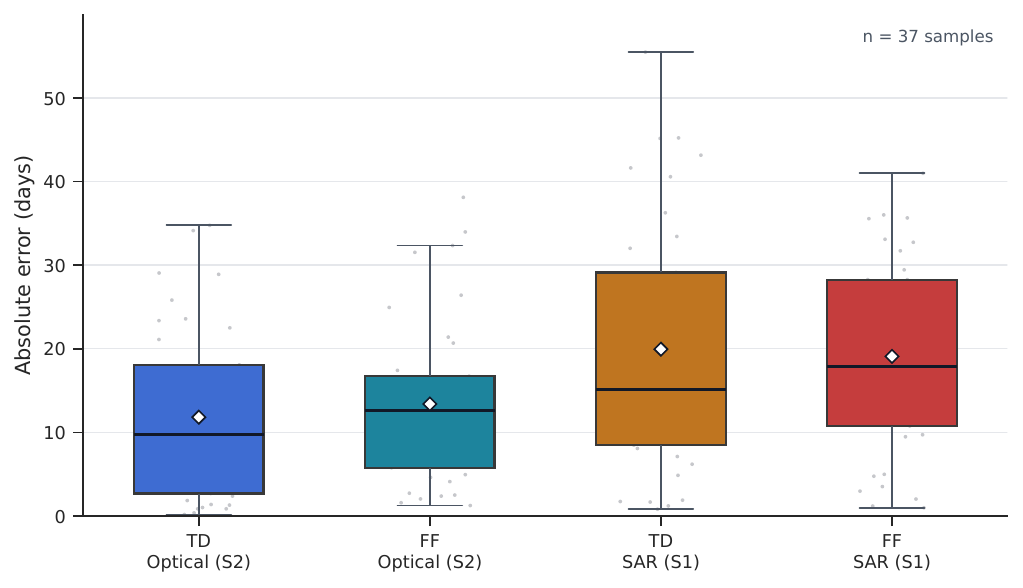}
  \caption{Distribution of absolute harvest-date prediction errors across the validation set of SICKLE. Results are shown for TD and FF considering both optical (S2) and SAR (S1) pretraining. Boxes represent the interquartile range, horizontal lines indicate medians, diamonds
  denote means, and points represent individual samples. Lower values indicate more accurate harvest-date predictions.}
  \label{fig:harvest_boxplot}
\end{figure}

\begin{figure*}[t]
\centering
\includegraphics[width=0.85\textwidth]{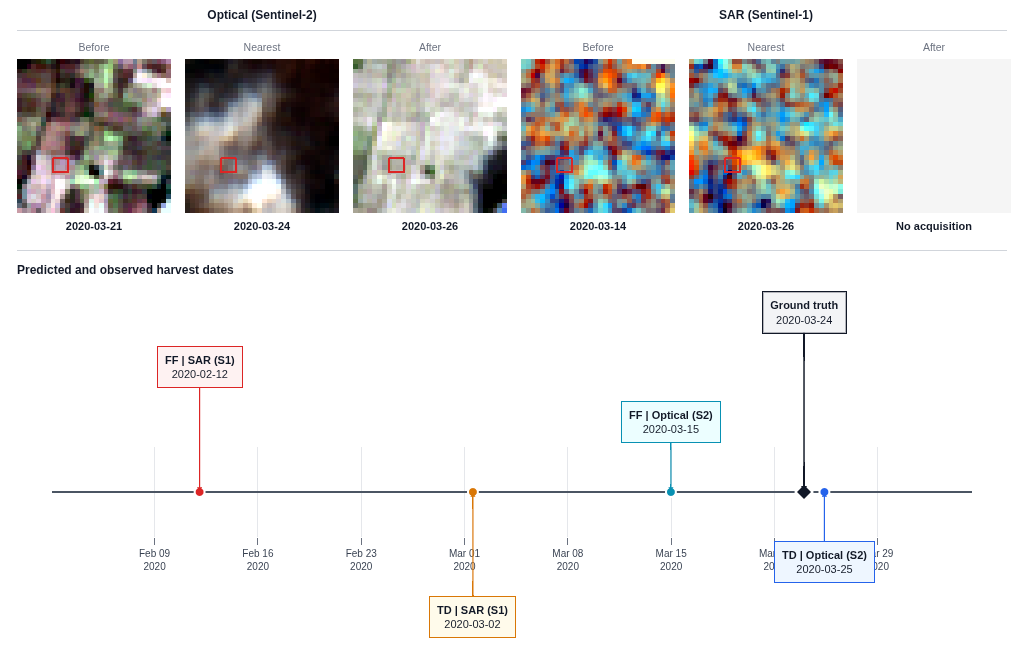}

\small (a)

\medskip
\includegraphics[width=0.85\textwidth]{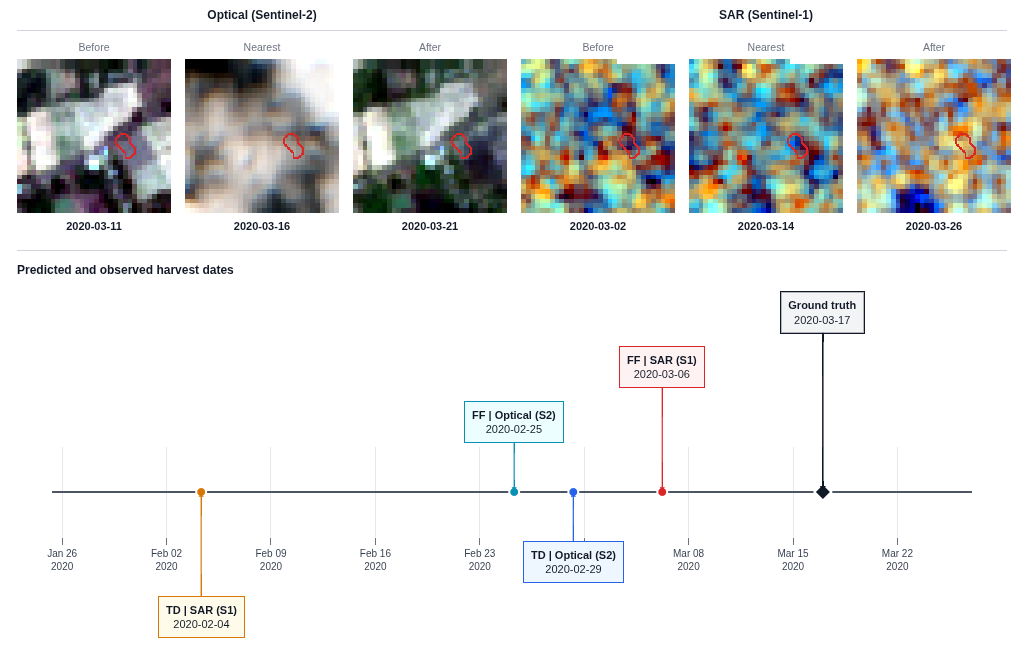}

\small (b) 

\caption{Some examples on harvest date prediction by TD \& FF pretrained on S1 and S2. Each row shows three S2 (RGB) and three S1 false-color acquisitions nearest to the ground truth harvest date, with farm boundaries delineated in red. S1 images are visualized as false-color composites (R: VV, G: VH, B: VH$-$VV) in log domain. Predicted harvest dates are shown for TD (S1), FF (S1), TD (S2), and FF (S2).
(a) TD (S2) achieves the lowest error at 1 day, relying on visible transition from 21/03/2020 to 26/03/2020 despite presence of cloud cover while SAR shows only a slight change.
(b) FF (S1) achieves the best prediction error (11 days) by correctly anticipating the harvest between the transition from dark blue (low VV, high VH and high VH-VV on 14/03/2020) to reddish-brown (intermediate VV, low VH and low VH-VV on 26/03/2020).
}
\label{fig:s1-s2-h12}
\end{figure*}

\begin{figure*}[t]
\centering
\includegraphics[width=0.85\textwidth]{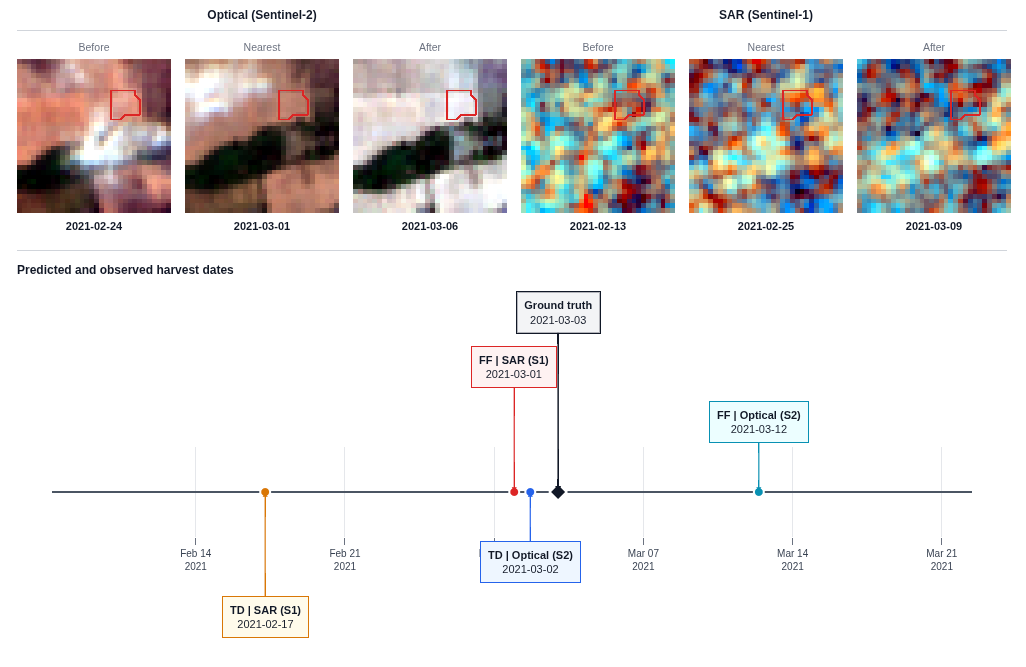}

\small (c)

\medskip
\includegraphics[width=0.85\textwidth]{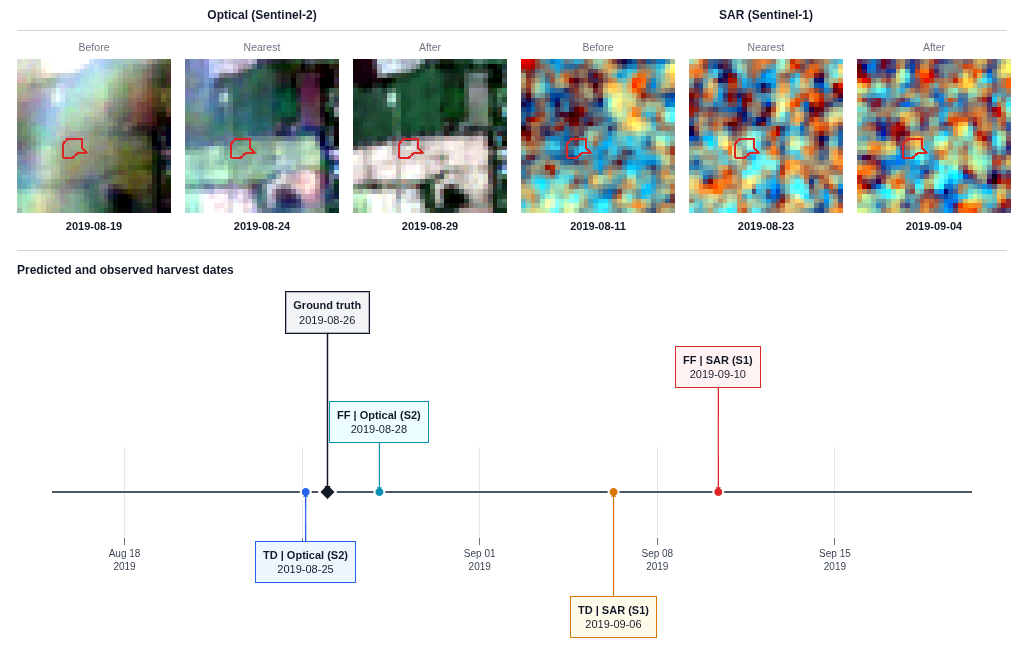}

\small (d)

\caption{Some examples on harvest date prediction by TD \& FF pretrained on S1 and S2 (continued from Fig~\ref{fig:s1-s2-h12}). Each row shows three S2 (RGB) and three S1 false-color acquisitions nearest to the ground truth harvest date, with farm boundaries delineated in red. S1 images are visualized as false-color composites (R: VV, G: VH, B: VH$-$VV) in log domain. Predicted harvest dates are shown for TD (S1), FF (S1), TD (S2), and FF (S2). 
(c) FF (S1) and TD (S2) achieve errors of 2 days and 1 day respectively. FF (S1) can correctly locate harvest within transition from 25/02/2021 to 09/03/2021 due to a more darker shade of red (high VV, low VH and low VH-VV), while TD (S2) can capture the stark visible transition from 01/03/2021 to 06/03/2021. 
(d) Both FF (S2) and TD (S2) achieve the lowest errors, at 2 days and 3 days respectively, due to the presence of strong transition in optical imagery (from 24/08/2019 to 29/08/2019).}
\label{fig:s1-s2-h34}
\end{figure*}

To understand the decline in harvest date prediction performance on S1, we compare absolute error across pretraining modality (S1 vs.\ S2) and pretext tasks TD \& FF in Fig~\ref{fig:harvest_boxplot}. Models pretrained on S2 achieve a lower median and mean error, suggesting that the optical modality provides a more reliable signal for harvest date prediction. However, the interquartile ranges of S1 and S2 overlap considerably, indicating that S1 can match or even outperform S2 on individual farms. Fig~\ref{fig:s1-s2-h12} and Fig~\ref{fig:s1-s2-h34} illustrate this overlap with qualitative examples across four farms. While TD (S2) achieves the lowest error in (a) and (c), FF (S1) outperforms all other configurations in (b), and FF (S2) and TD (S2) both outperform their S1 counterparts in (d), suggesting no single modality or pretext task is consistently superior across farms for harvest prediction.

\subsection{Comparison of S1 Pretraining}
\label{sec:harvest_viz}
\begin{figure*}[t]
\centering
\includegraphics[width=0.85\textwidth]{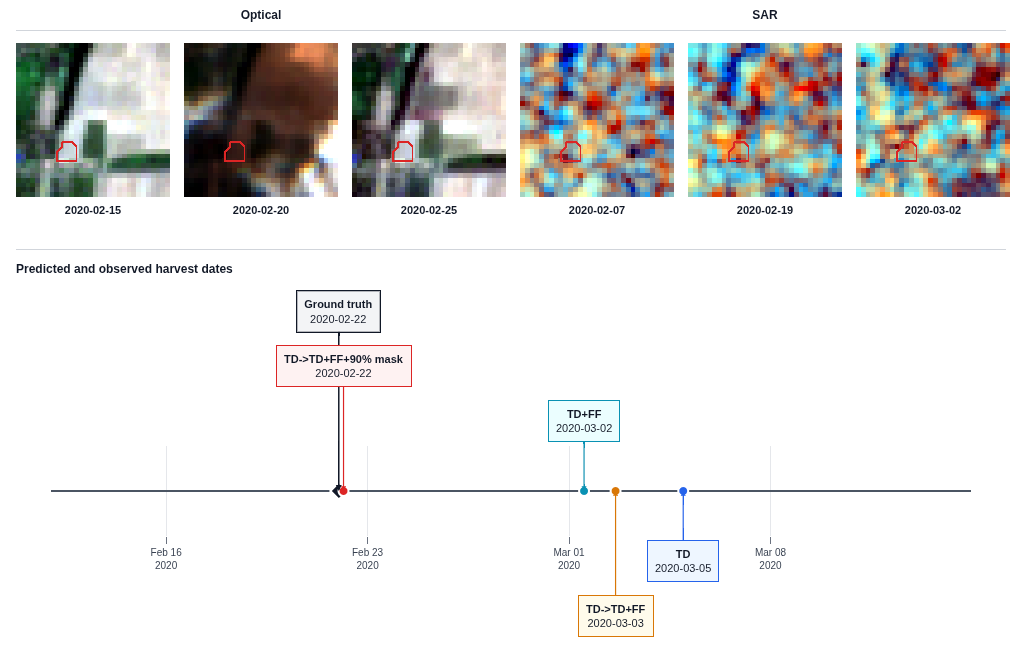}

\small (a)

\medskip
\includegraphics[width=0.85\textwidth]{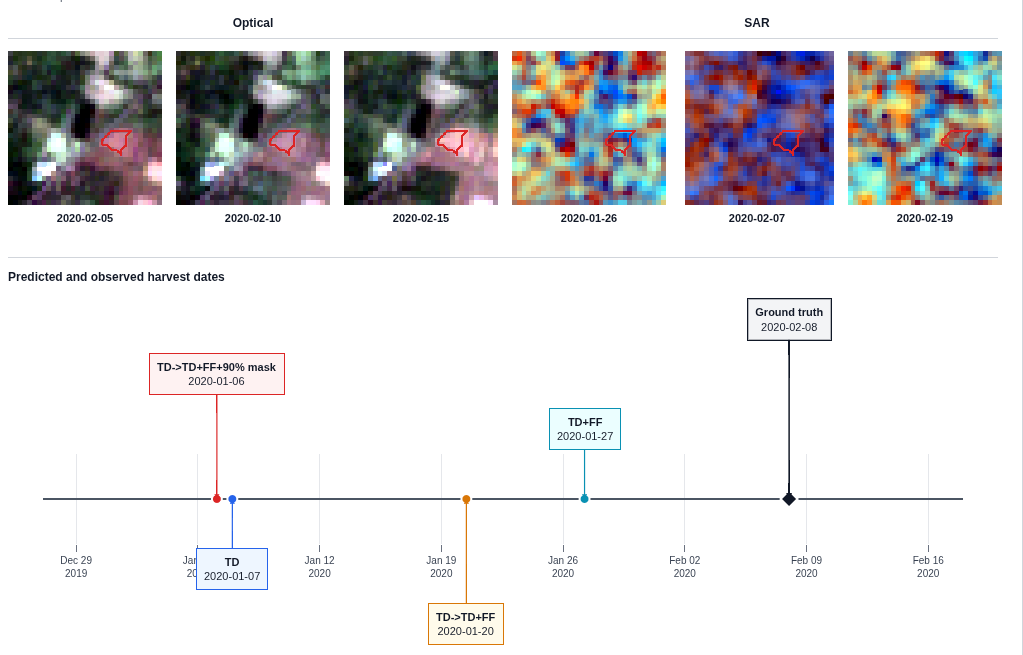}

\small (b) 

\caption{Some examples on harvest date prediction by our S1 pretrained models: TD, TD+FF, TD$\rightarrow$TD+FF and TD$\rightarrow$TD+FF+90\% masking. Each row shows three S2 (RGB) and three S1 false-color acquisitions nearest to the ground truth harvest date, with farm boundaries delineated in red. S1 images are visualized as false-color composites (R: VV, G: VH, B: VH$-$VV) in log domain. Predicted harvest dates are shown for TD, TD+FF, TD$\rightarrow$TD+FF and TD$\rightarrow$TD+FF+90\% masking. 
(a) TD$\rightarrow$TD+FF+90\% masking perfectly overlaps with ground truth, correctly anticipating a wider spread of red (high VV, low VH and low VH-VV) from 19/02/2020 to 02/03/2020 signifying harvest, while the rest of methods deviates by at least 9 days.
(b) All baselines deviate significantly with a least deviation of 12 days (FF+TD). None of them anticipated a harvest close to peak canopy (low VV, high VH and high VV-VH on 07/02/2020). 
}
\label{fig:h12}
\end{figure*}

\begin{figure*}[t]
\centering
\includegraphics[width=0.85\textwidth]{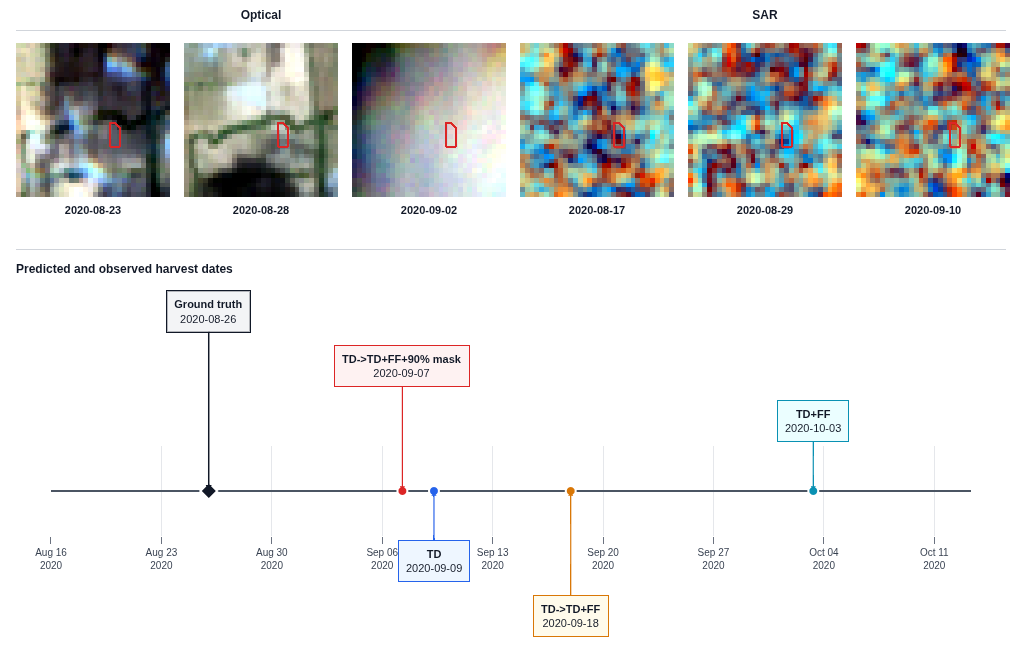}

\small (c)

\medskip
\includegraphics[width=0.85\textwidth]{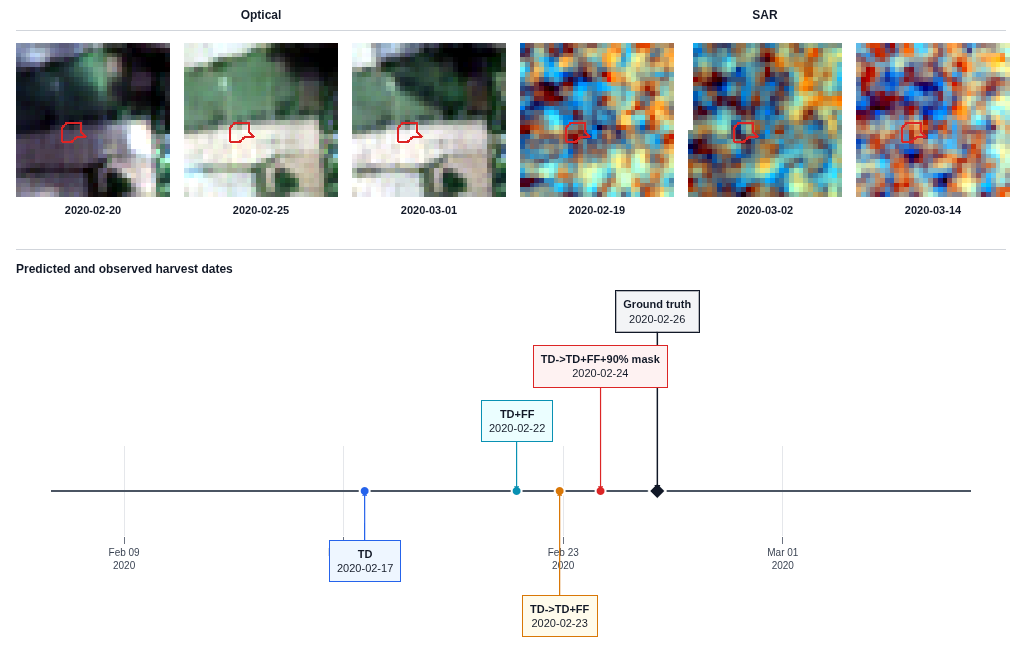}

\small (d) 

\caption{Some examples on harvest date prediction by our S1 pretrained models: TD, TD+FF, TD$\rightarrow$TD+FF and TD$\rightarrow$TD+FF+90\% masking (continued from Fig~\ref{fig:h12}). Each row shows three S2 (RGB) and three S1 false-color acquisitions nearest to the ground truth harvest date, with farm boundaries delineated in red. S1 images are visualized as false-color composites (R: VV, G: VH, B: VH$-$VV) in log domain. Predicted harvest dates are shown for TD, TD+FF, TD$\rightarrow$TD+FF and TD$\rightarrow$TD+FF+90\% masking. 
(c) Both TD and TD$\rightarrow$TD+FF achieve similar performance with a deviation of 12--13 days. None of them anticipates the harvest between two acquistions showing large VH and VH-VV scattering (17/08/2020 and 29/08/2020).
(d) TD$\rightarrow$TD+FF+90\% masking achieve the least error of 2 days followed by TD$\rightarrow$TD+FF (3 days) and TD+FF (4 days). All of them correctly captures the harvest event from 02/03/2020 to 14/03/2020 due to increase in VV, but decrease in VH and VH-VV.
}
\label{fig:h34}
\end{figure*}

Fig~\ref{fig:h12} and Fig~\ref{fig:h34} show some examples of harvest date prediction across four farms, comparing our S1 pretrained models: TD, TD+FF (MTL), TD$\rightarrow$TD+FF (curriculum), and TD$\rightarrow$TD+FF+90\% masking (ours). Our final pipeline: TD$\rightarrow$TD+FF+90\% masking, achieves the lowest harvest prediction error in 3 out of 4 cases (a, c and d).

\section{Implementation Details}
\label{sec:impl_details}
\subsection{Experimental Environment}
\label{subsec:experimental_environment}

Table~\ref{tab:env} describes the experimental environment for both pretraining and downstream evaluation. For all downstream experiments, we use RTX A5000 (24GB) with exception of SARATR-X \cite{li2025saratrxbuildingfoundationmodel} and SAR-W-MixMAE \cite{caglayan2026sar}, where we use an A100 (80GB) due to their high compute requirements.

\subsection{Pretraining \& Downstream Hyperparameters}
\label{sec:hparam}

Table~\ref{tab:pretrain_hparams} describes the hyperparameters for pretraining FF, TD, FF+TD, TD$\rightarrow$FF+TD, MAE, TD+Masking, FF+Masking, TD$\rightarrow$FF+TD+Masking, with curriculum learning experiments using hyperparameters according to the stage (e.g., TD$\rightarrow$TD+FF use TD's hyperparameters for Stage~I and FF+TD's hyperparameters for Stage~2).
Table~\ref{tab:downstream_hparams} describes the hyperparameters for downstream evaluation using DoFA \cite{xiong2024neuralplasticityinspiredmultimodalfoundation}, CopernicusFM \cite{wang2025unifiedcopernicusfoundationmodel}, TerraMind \cite{jakubik2025terramindlargescalegenerativemultimodality}, SAR-JEPA \cite{LI2024326}, SARATR-X \cite{li2025saratrxbuildingfoundationmodel}, SARMAE \cite{liu2026sarmae}, SAR-W-MixMAE \cite{caglayan2026sar}, UNet and ViT-S on the SICKLE benchmark.

\subsection{Model Architecture}
\label{sec:model_arch}
\subsubsection{Pretraining}
Table~\ref{tab:architecture_details}a describes the ViT-S \cite{dosovitskiy2021imageworth16x16words} backbone which is used by all pretraining methods. The pretext tasks require task-specific components, whose configurations are described in Table~\ref{tab:architecture_details}b. 
\subsubsection{Downstream}
For downstream evaluation on SICKLE, we follow \citet{Sani_2024_WACV} and \citet{gupta2026time2agri}, and use a SITS for prediction. To ingest SITS with a single-image backbone, we extract the backbone features at multiple layers (third, sixth, ninth and last layer for Small and Base variants of backbones) for each timestamp and perform a per-layer temporal max pooling. The resulting features are interpolated to form a feature pyramid and are ingested by the UperNet decoder for prediction, similar to \citet{xiong2024neuralplasticityinspiredmultimodalfoundation} and \citet{wang2025unifiedcopernicusfoundationmodel}. SARATR-X \cite{li2025saratrxbuildingfoundationmodel} and  SAR-W-MixMAE \cite{caglayan2026sar} do not perform interpolation since they natively output hierarichal features which can be directly used with UperNet.

\bibliography{aaai2027}


\end{document}